\documentclass[12pt,english]{article}
\usepackage[T1]{fontenc}
\usepackage[latin9]{inputenc}
\usepackage{geometry}
\usepackage{color}
\usepackage{amsmath}
\usepackage{amssymb}
\usepackage{graphicx}
\usepackage[authoryear]{natbib}

\makeatletter
\usepackage{enumitem}		% customizable list environments
\usepackage{hyperref}

\makeatother

\usepackage{babel}
\begin{document}
\title{Optimizing Rescoring Rules with Interpretable Representations of Long-Term
Information}
\author{Aaron Fisher}
\maketitle
\begin{abstract}
Analyzing temporal data (e.g., wearable device data) requires a decision
about how to combine information from the recent and distant past.
In the context of classifying sleep status from actigraphy, Webster's
rescoring rules offer one popular solution based on the long-term
patterns in the output of a moving-window model. Unfortunately, the
question of how to optimize rescoring rules for any given setting
has remained unsolved. 

To address this problem and expand the possible use cases of rescoring
rules, we propose rephrasing these rules in terms of epoch-specific
features. Our features take two general forms: (1) the \emph{time
lag} between now and the most recent {[}or closest upcoming{]} bout
of time spent in a given state, and (2) the \emph{length} of the most
recent {[}or closest upcoming{]} bout of time spent in a given state.
Given any initial moving window model, these features can be defined
recursively, allowing for straightforward optimization of rescoring
rules. Joint optimization of the moving window model and the subsequent
rescoring rules can also be implemented using gradient-based optimization
software, such as Tensorflow. Beyond binary classification problems
(e.g., sleep-wake), the same approach can be applied to summarize
long-term patterns for multi-state classification problems (e.g.,
sitting, walking, or stair climbing). We find that optimized rescoring
rules improve the performance of sleep-wake classifiers, achieving
accuracy comparable to that of certain neural network architectures.
\end{abstract}

\section{Introduction\label{sec:Introduction}}

An important question in temporal data analysis is how to weigh information
from the recent past against information from the distant past. Here,
we aim to inform this question by building on the framework of \emph{rescoring
rules}, \citep{Webster1982-cb}, a well-known method from the actigraphy
literature.

\begin{figure}
\begin{centering}
\includegraphics[width=0.95\columnwidth]{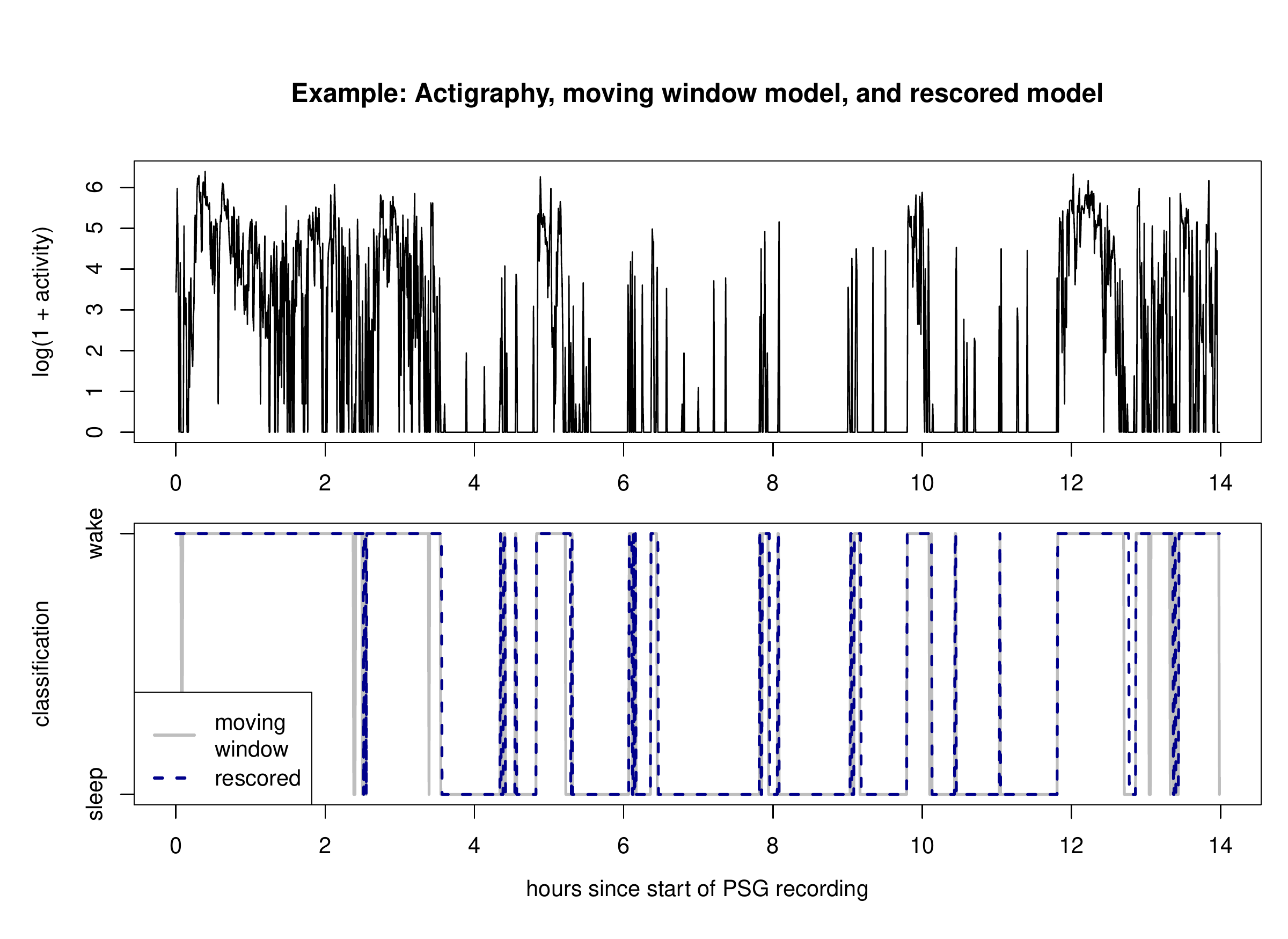}
\par\end{centering}
\caption{\label{fig:Example-Webster}Example application of Webster's rescoring
rules -- The top panel shows an example of activity over the course
of the night, taken from the MESA dataset (see Section \ref{sec:Performance-comparison-with}).
A simple moving window classifier was applied to this activity trajectory,
producing the sequence of sleep/wake classifications shown in the
bottom panel, in gray. After applying the moving window classifier,
Webster's rescoring rules were applied, producing the blue, dashed
line overlaid in the bottom panel. This rescoring procedure aims to
correct for epochs erroneously classified as sleep.}

\end{figure}

Numerous actigraphy studies have used moving window algorithms (MWAs)
to predict sleep status (\citealt{Webster1982-cb,Cole1992-ev,Sadeh1994-pg,Oakley1997-dy,Sazonov2004-pa};
see also the supplementary materials of \citealt{Palotti2019-qm}
for an especially cohesive summary). Beyond local information in the
moving window, \citealt{Webster1982-cb} proposed a post hoc series
of steps that can be applied to the output of a given MWA in order
to incorporate long-term activity patterns. These steps, known as
``Webster's rescoring rules'' for sleep-wake classification, are
widely popular, and are frequently referenced as a benchmark that
new methods can be compared against \citep{jean2000sleep,benson2004measurement,Palotti2019-qm,Haghayegh2019-jy,Haghayegh2020-wt}.
In their most general form, Webster's rescoring rules can be written
as follows, with tuning parameters (constants) $a,b,c$ and $d$.
\begin{description}
\item [{Rule$\,\,$1:}] After at least $a$ continuous minutes scored by
the MWA as wake, identify the next $b$ minutes and \emph{rescore}
these minutes to wake.
\item [{Rule$\,\,$2:}] If any bout lasting $c$ minutes or less has been
scored by the MWA as sleep, and is surrounded by at least $d$ minutes
(before and after) scored by the MWA as wake, \emph{rescore} this
bout to wake.
\end{description}
The first rule reflects the idea that inactivity onset usually precedes
sleep onset by several minutes. The second rule reflects the idea
that brief sedentary periods do not necessarily indicate sleep, especially
if they are surrounded by long periods of activity. \citealt{Webster1982-cb}
suggest applying several different versions of each rule simultaneously,
setting $(a,b)$ to $(4,1)$, $(10,3)$, and $(15,4)$; and setting
$(c,d)$ to $(6,10)$ and $(10,20)$. The resulting rules are illustrated
in Figures \ref{fig:Example-Webster}.

By adjusting for long term patterns, these post hoc rules can make
the accuracy of simple moving window models closer to that of recurrent
neural network models (RNNs, \citealp{Palotti2019-qm}). This improvement
is intuitive, as RNNs often aim to find an optimal representation
of long term patterns, after applying an initial moving window (i.e.,
convolutional) step. An advantage of Webster's rules is that their
interpretability helps users to understand when rescoring might not
be appropriate, while black-box RNN rules can produce failures that
are more difficult to identify. 

Several disadvantages of Webster's rules remain though. First, these
rules do not produce continuous predicted class probabilities, only
binary classifications. More importantly, these rules appear to have
been derived heuristically through trial and error, rather than from
being framed as a formal optimization problem. Some authors have presented
alternative formulations of rescoring rules, but these appear to be
heuristic as well \citep{qian2015wake,kripke2010wrist}. Indeed, because
of the somewhat complex structure of Webster's rules, it is not immediately
obvious how the constants $a,b,c$ and $d$ should be formally optimized,
or how they should be recalibrated for new populations (\citealp{Ludtke2021-xu};
see also \citealp{heglum2021distinguishing}). These challenges deepen
if we consider joint optimization of the moving window algorithm and
the rescoring steps, rather than sequential optimization. Perhaps
for these reasons, several modern papers on sleep-wake classification
apply Webster's rules as purely ``off-the-shelf,'' with no calibration
(\citealp{tilmanne2009algorithms,Palotti2019-qm,Liu2020-ly}).

These questions are the inspiration for our work. Namely, we demonstrate
how rescoring rules can be optimized and/or calibrated by rephrasing
these rules in terms of epoch-level features, such as the length of
the most recent bout in a given state. Section \ref{sec:rescore-features}
introduces these features, Section \ref{sec:Improving-Webster's-rescoring}
discusses optimization methods, and Section \ref{sec:Performance-comparison-with}
studies the performance of our approach in the Multi-Ethnic Study
of Atherosclerosis (MESA) sleep study dataset \citep{Chen2015-fz,Zhang2018-kg}.
We find that optimizing rules improves the performance of moving window
models, although the difference is less pronounced for models with
larger windows. We close with a discussion. In particular, we note
that, while our proposed methods are motivated by sleep studies, they
can also be applied to general, multi-state classification problems.

\section{Rewriting rescoring rules with epoch-level features\label{sec:rescore-features}}

In this section, as a first step towards choosing optimal parameters
for rescoring rules (e.g., parameters $a,b,c$ and $d$ in Section
\ref{sec:Introduction}), we will define a set of epoch-level, recursive
features. We will then illustrate how Webster's rescoring rules can
be reexpressed in terms of those features.

Let $\mathbf{X}=(X_{1},\dots,X_{T})$, where $X_{t}$ is a sleep study
participant's activity summary at time $t$ during the night. Here,
$t$ ranges from $1$ to $T$, with each value denoting one epoch.
For simplicity of presentation, we omit an index for participants
in a dataset, and instead focus on sleep/wake transitions for a single
participant on a single night. Let $\mathbf{Y}=(Y_{1},\dots Y_{T})$,
where $Y_{t}=1$ indicates wake at time $t$ and $Y_{t}=0$ indicates
sleep at time $t$. For a given moving window algorithm applied to
activity, let $\hat{\pi}=(\hat{\pi}_{1},\dots,\hat{\pi}_{T})$, where
$\hat{\pi}_{t}$ is the estimate of $P(Y_{t}=1)$ produced by that
algorithm. Let $\mathbf{W}=(W_{1},\dots W_{T})$, where $W_{t}\in\{0,1\}$
is a thresholded version of $\hat{\pi}_{t}$ and $W_{t}=1$ indicates
a prediction of wake. 

\begin{figure}
\begin{centering}
\includegraphics[width=1.02\columnwidth]{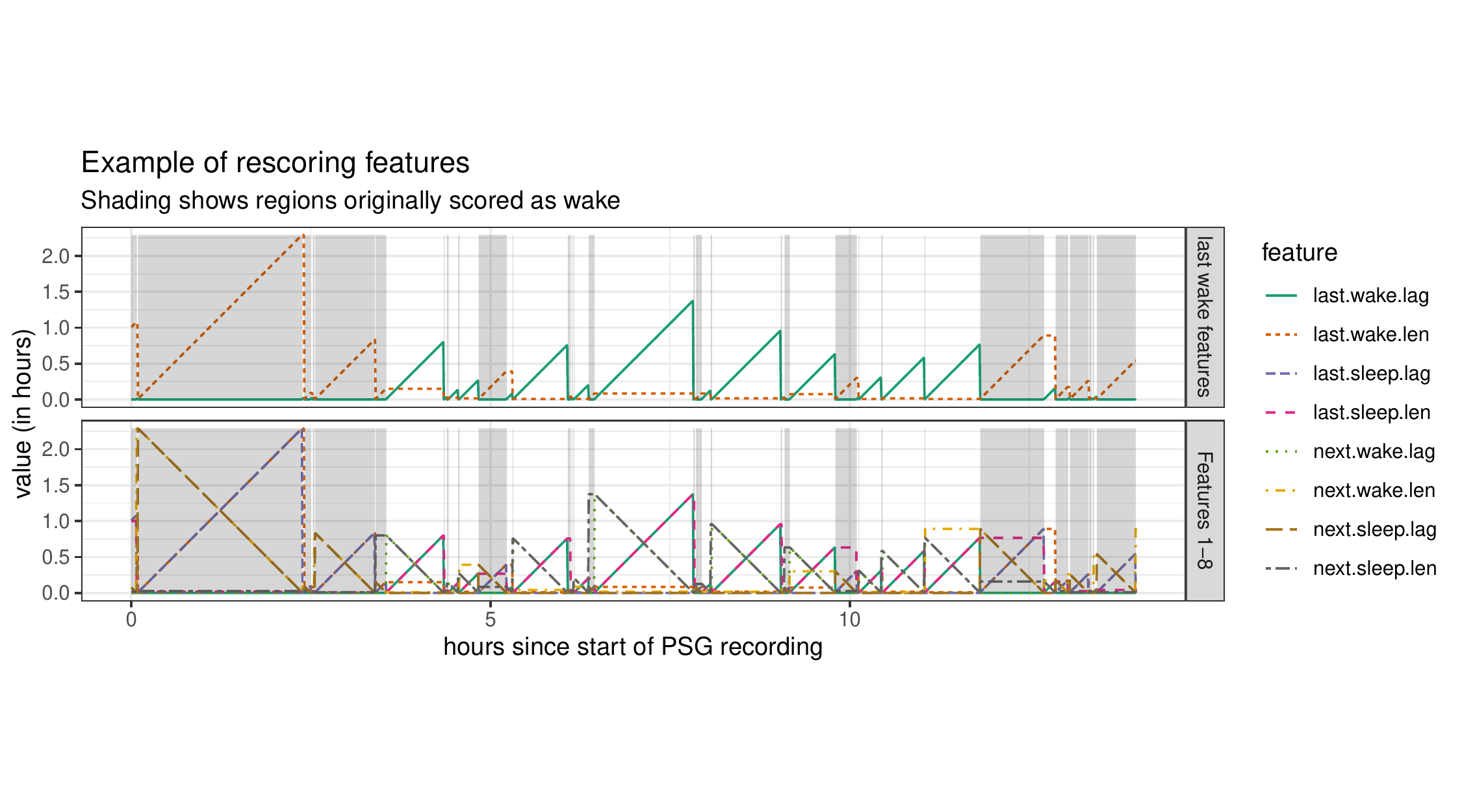}
\par\end{centering}
\caption{\label{fig:Example-of-rescoring}Example of rescoring features --
Here we show the rescoring features, computed as a function of the
classification sequence shown in the bottom panel of Figure \ref{fig:Webster's-rescoring-rules}
($W_{t}$, gray line). The original classification from the moving
window model $(W_{t}$) is shown here via shading, with gray corresponding
to regions classified as wake. The first panel shows two rescoring
features, $\text{last}_{\text{lag}}^{\text{wake}}(\mathbf{W},t)$
and $\text{last}_{\text{len}}^{\text{wake}}(\mathbf{W},t)$. These
two functions represent the core operations of our proposed procedure
-- all other values in $(l(\mathbf{W},t),n(\mathbf{W},t))$ can be
found reversing the time index (x-axis) and/or by flipping the original
score (shading). The second panel shows the 8 features in $(l(\mathbf{W},t),n(\mathbf{W},t))$
together.}
\end{figure}

In order to capture long-term patterns in sleep\textcolor{black}{,
we define the several features for each epoch, indexed by $t$. These
features have two general forms: (1) the }\textcolor{black}{\emph{time
lag}}\textcolor{black}{{} between $t$ and the most recent {[}or closest
upcoming{]} bout of time spent in a given state, and (2) the }\textcolor{black}{\emph{length}}\textcolor{black}{{}
of the most recent {[}or closest upcoming{]} bout of time spent in
a given state. More explicitly, we define these features as follows. }
\begin{enumerate}
\item $\text{last}_{\text{lag}}^{\text{wake}}(\mathbf{Y},t)$: the time
between $t$ and the most recent epoch $t'\leq t$ for which $Y_{t'}=1$.
\item $\text{last}_{\text{lag}}^{\text{sleep}}(\mathbf{Y},t)$: the time
between $t$ and the most recent epoch $t'\leq t$ for which $Y_{t'}=0$.
\item $\text{last}_{\text{len}}^{\text{wake}}(\mathbf{Y},t)$: if $Y_{t}=0$,
this feature returns the length of the most recent wake bout. If $Y_{t}=1$,
it returns the cumulative length of the current wake bout.
\item $\text{last}_{\text{len}}^{\text{sleep}}(\mathbf{Y},t)$: if $Y_{t}=1$,
this feature returns the length of the most recent sleep bout. If
$Y_{t}=0$, it returns the cumulative length of the current sleep
bout.
\item $\text{next}_{\text{lag}}^{\text{wake}}(\mathbf{Y},t)$: the time
between $t$ and the closest upcoming epoch $t'\geq t$ for which
$Y_{t'}=1$.
\item $\text{next}_{\text{lag}}^{\text{sleep}}(\mathbf{Y},t)$: the time
between $t$ and the closest upcoming epoch $t'\geq t$ for which
$Y_{t'}=0$.
\item $\text{next}_{\text{len}}^{\text{wake}}(\mathbf{Y},t)$: if $Y_{t}=0$,
this feature returns the length of the closest upcoming wake bout.
If $Y_{t}=1$, it returns the remaining length of the current wake
bout.
\item $\text{next}_{\text{len}}^{\text{sleep}}(\mathbf{Y},t)$: if $Y_{t}=1$,
this feature returns the length of the closest upcoming t sleep bout.
If $Y_{t}=0$, it returns the remaining length of the current sleep
bout.
\end{enumerate}
\textcolor{black}{Figure \ref{fig:Example-of-rescoring} ill}ustrates
how each of the above features transform their input, $\mathbf{Y}$.
We refer to the first four features collectively as $l(\mathbf{Y},t)$,
and refer to the last four features as $n(\mathbf{Y},t)$. The second
set of features, $n(\mathbf{Y},t)$, can also be attained by reversing
the time index of $l(\mathbf{Y},t)$. That is, if $\mathbf{Y}_{[T:1]}=(Y_{T},Y_{T-1},\dots Y_{2},Y_{1})$
is a time-reversed version of $\mathbf{Y}$, such that $(\mathbf{Y}_{[T:1]})_{t}=Y_{T-t+1}$,
then $n(\mathbf{Y},t)=l(\mathbf{Y}_{[T:1]},T-t+1)$.

Additionally, given $l(\mathbf{Y},t)$ and $n(\text{\textbf{Y}},t)$,
we can compute summaries of the sleep patterns \emph{surrounding}
any given epoch. In particular, we consider the following aggregate
features.
\begin{enumerate}[resume]
\item $\text{current.total}_{\text{len}}^{\text{sleep}}(\mathbf{W},t)$:
the length of the current sleep bout, equal to $\text{last}_{\text{lag}}^{\text{wake}}(\mathbf{W},t)+\text{next}_{\text{lag}}^{\text{wake}}(\mathbf{W},t).$
\item $\text{\text{current.total}}_{\text{len}}^{\text{wake}}(\mathbf{W},t):$
the length of the current wake bout, equal to $\text{last}_{\text{lag}}^{\text{sleep}}(\mathbf{W},t)+\text{next}_{\text{lag}}^{\text{sleep}}(\mathbf{W},t)$.
\item $\text{min.bordering}_{\text{len}}^{\text{sleep}}(\mathbf{W},t)$:
the length of the smallest bordering sleep bout, equal to $\min\left(\text{\text{last}}_{\text{len}}^{\text{sleep}}(\mathbf{W},t),\text{\text{next}}_{\text{len}}^{\text{sleep}}(\mathbf{W},t)\right)$.
\item $\text{min.bordering}_{\text{len}}^{\text{wake}}(\mathbf{W},t)$:
the length of the smallest bordering wake bout, equal to $\min\left(\text{\text{last}}_{\text{len}}^{\text{wake}}(\mathbf{W},t),\text{\text{next}}_{\text{len}}^{\text{wake}}(\mathbf{W},t)\right)$.
\end{enumerate}
We refer to the above four features as $c(\mathbf{Y},t)$.

An important property of these 12 features is that they can all be
defined recursively (shown in Appendix \ref{sec:Recursive-definitions-for}).
Given $l(\mathbf{Y},t-1)$, the features $l(\mathbf{Y},t)$ depend
only on $Y_{t}$. Similarly, given $n(\mathbf{Y},t+1)$, the features
$n(\mathbf{Y},t)$ depend only on $Y_{t}$. Further, given $l(\mathbf{Y},t-1)$
and $n(\mathbf{Y},t+1)$ the features $l(\mathbf{Y},t)$ and $n(\mathbf{Y},t)$
are both\emph{ linear} in $Y_{t}$ (see Appendix \ref{sec:Recursive-definitions-for}).
From here, $c(\mathbf{Y},t)$ is determined immediately from $(l(\mathbf{Y},t),n(\mathbf{Y},t)$).
This recursive structure will prove useful in the next section, and
is illustrated in Figure \ref{fig:structure}. 

In practice, none of the above features are known for unlabeled data,
but they can be approximated by plugging in $\mathbf{W}$ for $\mathbf{Y}$
to yield the vector of \emph{rescoring features} $r(\mathbf{W},t):=(W_{t},l(\mathbf{W},t),n(\mathbf{W},t),c(\mathbf{W},t))$.
Since $l(\mathbf{W},1)$ and $n(\mathbf{W},T)$ depend on information
outside of the time period in which participants are observed, we
require user-supplied tuning parameters $b_{1},b_{T}\in\mathbb{R}_{\geq0}^{4}$,
and set these border values to be $l(\mathbf{W},1)=b_{1}$ and $n(\mathbf{W},T)=b_{T}$.

\begin{figure}
\begin{centering}
\includegraphics[width=0.95\columnwidth]{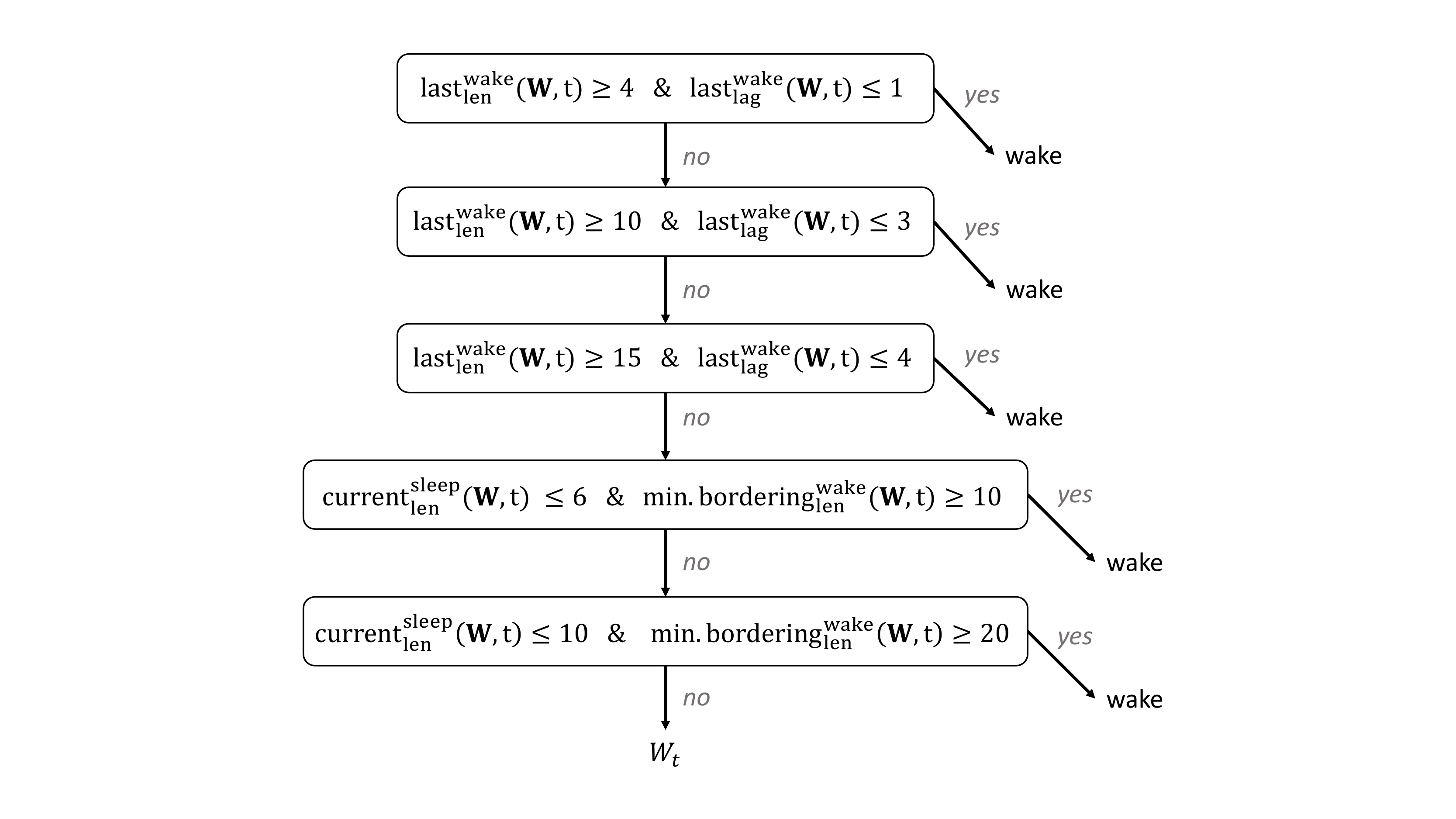}
\par\end{centering}
\caption{\label{fig:Webster's-rescoring-rules}Webster's rescoring rules rewritten
in terms of the features from Section \ref{sec:rescore-features}.
This tree returns the rescored prediction of sleep/wake status at
epoch $t$.}
\end{figure}

Finally, with these features in mind, we can see that the two general
forms of Webster's rescoring rules can be rewritten as follows. 
\begin{description}
\item [{Rule$\,\,$1:}] if $\text{last}_{\text{len}}^{\text{wake}}(\mathbf{W},t)\geq a$
and $\text{last}_{\text{lag}}^{\text{wake}}(\mathbf{W},t)\leq b$,
then rescore the $t^{th}$ epoch as wake.
\item [{Rule$\,\,$2:}] if $\text{current.total}_{\text{len}}^{\text{sleep}}(\mathbf{W},t)\leq c$
and $\text{min.bordering}_{\text{len}}^{\text{wake}}(\mathbf{W},t)\geq d$,
then rescore the $t^{th}$ epoch as wake.
\end{description}
Any epoch that does not meet e\textcolor{black}{ither of these rules
for any of the allowed values of $a,b,c$ and $d$ retains its original
score}\textbf{\textcolor{black}{{} ($W_{t}$}}\textcolor{black}{). In
other words, }Webster's rules form a decision tree with $r(\mathbf{W},t)$
as input (\textcolor{black}{see Figure \ref{fig:Webster's-rescoring-rules}}). 

\section{Optimizing rescoring rules\label{sec:Improving-Webster's-rescoring}}

As Webster's rescoring rules are a heuristically chosen function of
$r(\mathbf{W},t)$, a natural next step is to search for an optimal
function of $r(\mathbf{W},t)$. This is especially straightforward
if we treat the original scores $\mathbf{W}$ as fixed, in which case
we can apply any off-the-shelf training algorithm to find a classifier
of $Y_{t}$ based on $r(\mathbf{W},t)$. One caveat is that, since
$c(\mathbf{W},t)$ includes linear combinations of $l(\mathbf{W},t)$
and $n(\mathbf{W},t)$, it cannot be included alongside $l(\mathbf{W},t)$
and $n(\mathbf{W},t)$ in models that require linearly independent
covariates (e.g., logistic regression models), at least not without
first applying a log or other nonlinear transformation.

We can also choose to include a continuous version of $r\left(\mathbf{W},t\right)$
in our classifier. One straightforward implementation is to replace
$W_{t}$ with the unthresholded probabilities $\hat{\pi}$, forming
$r(\hat{\pi},t)=\left(\hat{\pi},l(\hat{\pi},t),n(\hat{\pi},t),c(\hat{\pi},t)\right)$.
This implementation can  be motivated by the working model assumption
that, given \textbf{$\mathbf{X}$}, the wake indicators $(Y_{1},\dots,Y_{T})$
are independently distributed Bernoulli variables with $P(Y_{t}=1|\mathbf{X})=\hat{\pi}_{t}$.
Since each element of $\left(l(\hat{\pi},t),n(\hat{\pi},t)\right)$
is linear in $Y_{t}$ given its neighbors, it follows from this working
model that $E\left[l(\mathbf{Y},t)|\mathbf{X}\right]=l(\hat{\pi},t)$
and $E\left[n(\mathbf{Y},t)|\mathbf{X}\right]=n(\hat{\pi},t)$ (see
Appendix \ref{sec:Expectation-of}). 

\begin{figure}
\begin{centering}
\includegraphics[width=0.96\columnwidth]{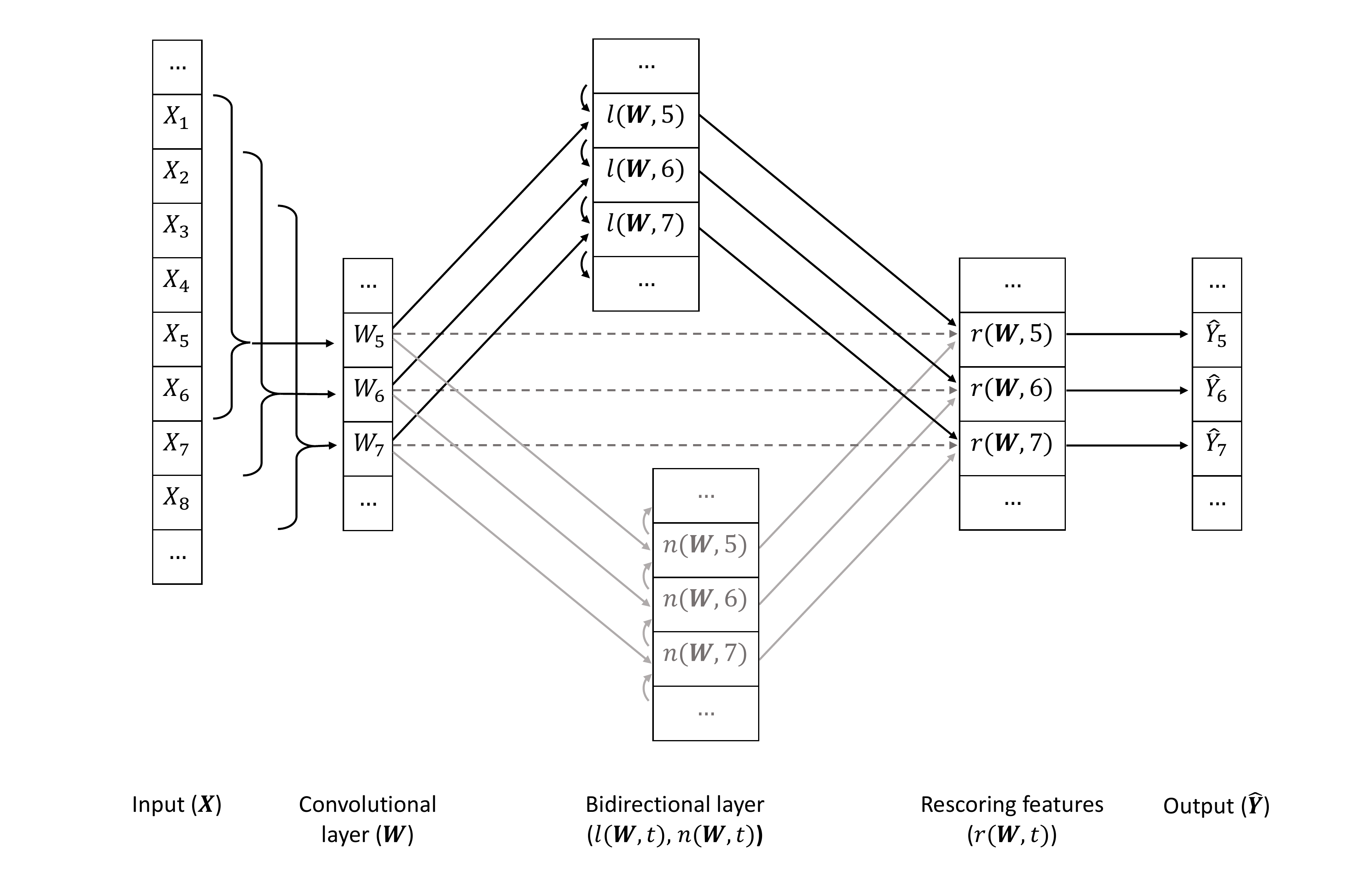}
\par\end{centering}
\caption{\label{fig:structure}Rescoring pipeline -- Here we show the general
structure of Webster's rescoring rules, where the moving window model
output $(W_{t})$ is treated as fixed. For the purposes of this illustration,
the input to the moving window model is $(X_{t-4},X_{t-3},X_{t-2},X_{t-1},X_{t},X_{t+1})$.
Each box refers to an epoch-level variable, or feature vector. Arrows
indicate the inputs for each feature. Webster's rescoring rules can
be interpreted as one way to transform from the penultimate layer
$r(\hat{\pi}_{t},t)$ to the prediction \textbf{$\hat{Y}_{t}$}. The
sequential optimization approach in Section \ref{sec:Improving-Webster's-rescoring}
can be attained by optimizing this transformation. The joint optimization
approach in Section \ref{sec:Improving-Webster's-rescoring} can be
attained by replacing each thresholded $W_{t}$ with a continuous
probability, and jointly optimizing both the convolutional layer and
the transformation from rescoring features to final predictions, while
fixing the intervening layers based on intuitive patterns suggested
by \citet{Webster1982-cb}. }
\end{figure}

From here, we can jointly optimize the continuous features $\hat{\pi}_{t}$
and the model fit to the features $r\left(\hat{\pi}_{t},t\right)$
using gradient-based optimization software, such as PyTorch or Tensorflow
\citep{tensorflow2015-whitepaper,paszke2017automatic_pytorch}. For
example, consider the multi-layer model
\begin{equation}
P(Y_{t}=1|\mathbf{X})=\text{logit}^{-1}\left[\alpha_{1}+\left(\text{logit}(\hat{\pi}_{t}),\log\left\{ 1+\left(l\left(\hat{\pi},t\right),n\left(\hat{\pi},t\right),c\left(\hat{\pi},t\right)\right)\right\} \right)^{\top}\beta_{1}\right],\label{eq:nn1}
\end{equation}
where
\begin{equation}
\ensuremath{\hat{\pi}}_{t}=\text{logit}^{-1}\left(\alpha_{0}+\left[\begin{array}{cccc}
X_{t+a} & X_{t+a+1} & \dots & X_{t+b}\end{array}\right]\beta_{0}\right).\label{eq:nn2}
\end{equation}
Above, $(\alpha_{0},\alpha_{1},\beta_{0},\beta_{1})$ are model parameters,
and $(a,b)$ are tuning parameters determining the size of a moving
window, chosen to satisfy $a\leq0\leq b$. Under this model, the likelihood
of $Y_{t}$ given $\mathbf{X}$ is a differentiable (almost everywhere)
function of $(\alpha_{0},\alpha_{1},\beta_{0},\beta_{1})$, and so
maximum likelihood estimates can be attained using gradient-based
methods. While the requirement that we use the continuous quantity
$\hat{\pi}$ (rather than $\mathbf{W}$) may appear restrictive, we
note that $\hat{\pi}$ can be made arbitrarily close to a binary variable
by increasing the scale of $(\alpha_{0},\beta_{0})$. For our implementation,
we train the model in Eqs (\ref{eq:nn1})-(\ref{eq:nn2}) using the
r Keras package \citep{chollet2017kerasR}, with a customized recurrent
layer to represent $r\left(\hat{\pi},t\right)$. The structure of
the resulting network is shown in Figure \ref{fig:structure}.

\section{Performance comparison with MESA dataset\label{sec:Performance-comparison-with}}

In this section, we compare the performance of standard MWAs, MWAs
with off-the-shelf Webster's rescoring rules, and MWAs with calibrated
rescoring rules. In calibrating these rules, we apply both sequential
and joint optimization approaches. We evaluate these methods using
the commercial use dataset from the Multi-Ethnic Study of Atherosclerosis
(MESA), a longitudinal study of cardiovascular disease (\citealp{Chen2015-fz,Zhang2018-kg};
see also \citealp{Palotti2019-qm}). One night of concurrent actigraphy
and scored polysomnography (PSG) was recorded for many of the participants
in this study. After filtering to participants with at least 5 hours
of contiguously measured actigraphy and PSG, our dataset included
1685 individuals. 

As a baseline prediction model, we fit logistic regressions to predict
sleep state $Y_{t}$ using the variables $(X_{t+a},X_{t+a+1},X_{t+a+2},\dots X_{t+b})$
as input (referred to as ``GLM-window''). We considered three window
sizes, setting $[a,b]$ equal to $[-5,2]$, $[-10,5]$ and $[-30,20]$.
For each model, we also applied ``off-the-shelf'' Webster's rescoring
rules (referred to as ``Webster'').

For sequentially optimized rescoring rules, we applied the transformations
in Section \ref{sec:rescore-features} to the output of the three
logistic regression models above. For each window size, we fit a second,
post hoc logistic regression to predict sleep state $Y_{t}$ using
$\text{logit}(\hat{\pi})$, $\log\{1+l(\hat{\pi},t)\}$, $\log\{1+n(\hat{\pi},t)\}$,
and $\log\{1+c(\hat{\pi},t)\}$ as input (referred to as ``GLM-continuous''),
where $\hat{\pi}$ is the vector of predicted wake probabilities produced
by the ``GLM-window'' method. We also fit logistic regressions taking
($\hat{\pi},l(\mathbf{W},t),n(\mathbf{W},t),c(\mathbf{W},t)$) as
input, using 0.5 as the threshold for $\mathbf{W}$ (``GLM-binary'').

To jointly optimize the rescoring rules, we used the r Keras implementation
described in Section \ref{sec:Improving-Webster's-rescoring} (``rescore-NN'').
As above, we considered three window sizes: $[-5,2]$, $[-10,5]$,
and $[-30,20]$. We initialized each model using the coefficients
from the ``GLM-window'' and ``GLM-continuous'' methods. After
initialization, we trained each model with a batch size of 100, and
20 epochs. 

As a benchmark representing more complex prediction models, we compared
against neural networks with three layers: (1) a 1-dimensional convolutional
layer taking activity as input; (2) a bi-directional, long short-term
memory (LSTM) layer taking the output of Layer 1 as input; and (3)
a linear layer taking the output of both Layer 1 and Layer 2 as input
(analogous to Figure \ref{fig:structure}). We implemented this structure
again using the r Keras package \citep{chollet2017kerasR}, and considered
two configurations, referred to as ``LSTM-1-6'' and ``LSTM-10-30.''
The first configuration, ``LSTM-1-6,'' was meant to mimic the structure
and initialization procedure of ``rescore-NN,'' with a single convolutional
filter and 6 hidden variables in each direction of the LSTM layer
(12 total). We initialized Layer 1's weights based on the ``GLM-window''
model, and initialized Layer 3's weights to be 1 for the Layer 1 output,
and zero elsewhere. Thus, at initialization, ``LSTM-1-6'' produces
predictions identical to ``GLM-window.'' The second configuration,
``LSTM-10-30,'' allowed for a more complex structure, with 10 filters
in the convolutional layer and 30 hidden variables in each direction
of the LSTM layer. As above, both configurations were fit using windows
of $[-5,2]$, $[-10,5]$ and $[-30,20]$ for the convolutional layer.
The batch size and number of epochs were again set to 100 and 20 respectively.

We evaluated performance using 5-fold cross-validation, computing
Receiver Operating Characteristic (ROC) curves for each prediction
model. To attain ROC curves for Webster's rules, varied the threshold
used to define $\mathbf{W}$. 

\begin{figure}
\begin{centering}
\includegraphics[width=1\columnwidth]{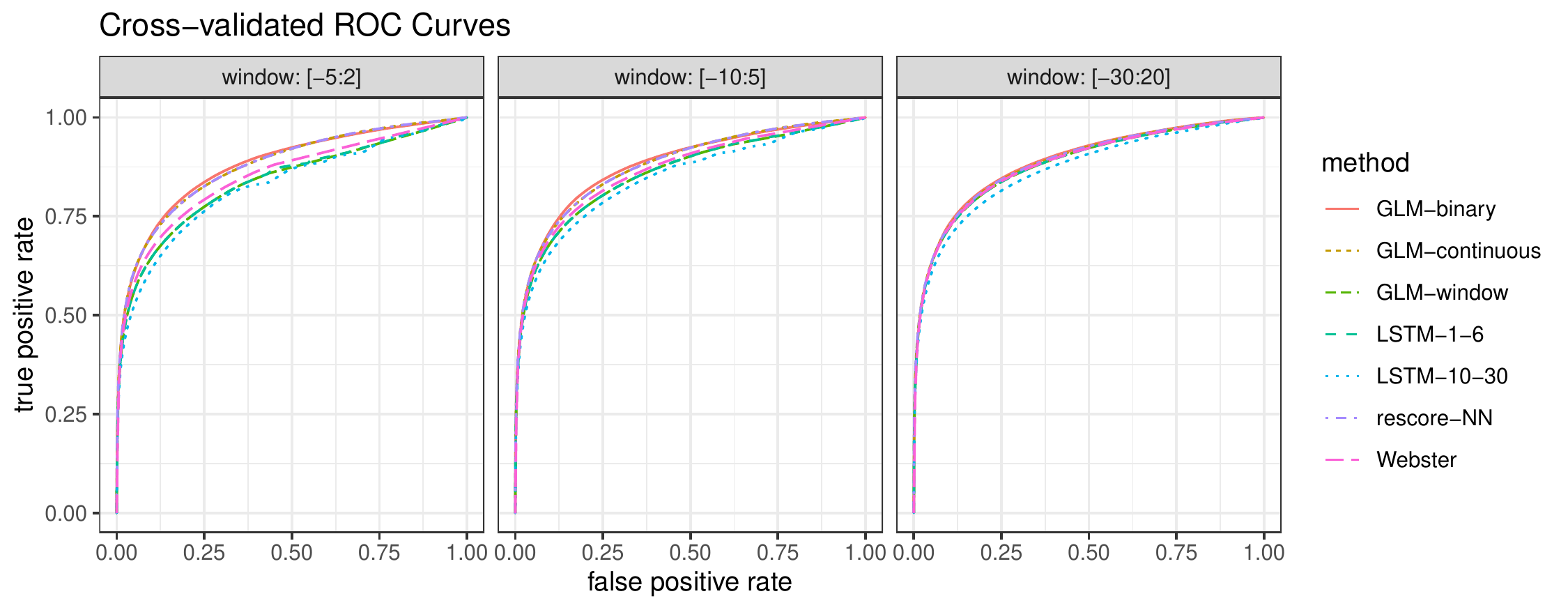}
\par\end{centering}
\caption{\label{fig:ROC}Receiver Operating Characteristic ROC curves for each
model.}
\end{figure}

Figure \ref{fig:ROC} shows ROC curves from our analysis, and Table
\ref{tab:AUC} shows the area under each curve (AUC). Webster's rules
gave a small performance improvement over our simplest model, ``GLM-window.''
Sequentially optimizing the rescoring rules (``GLM-continuous''
and ``GLM-binary'') added another small improvement over Webster's
rules. Jointly optimizing the moving window weights and the rescoring
rules (``rescore-NN'') had a negligible effect on performance, relative
to ``GLM-continuous.'' Our optimized rescoring methods outperformed
the LSTM models implementations described above. That said, this result
does not preclude the possibility of other neural network architectures
generating better performance from LSTM layers. The differences between
the above methods were also diminished when longer windows were used.
\begin{table}
\begin{centering}
% latex table generated in R 4.0.5 by xtable 1.8-4 package % Wed Apr 28 09:08:02 2021 
\begin{tabular}{lrrr}
  \hline
   & Window & & \\
 Method & [-5:2] & [-10:5] & [-30:20] \\ 
  \hline
  \bf{GLM-binary} & \bf{0.879} & \bf{0.882} & \bf{0.888} \\ 
  \bf{GLM-continuous} & \bf{0.877} & \bf{0.880} & \bf{0.886} \\ 
  GLM-window & 0.836 & 0.860 & 0.882 \\ 
  LSTM-1-6 & 0.837 & 0.860 & 0.882 \\ 
  LSTM-10-30 & 0.827 & 0.846 & 0.868 \\ 
  \bf{rescore-NN} & \bf{0.877} & \bf{0.880} & \bf{0.886} \\ 
  Webster & 0.850 & 0.867 & 0.882 \\ 
   \hline
\end{tabular} 
\caption{\label{tab:AUC}Areas under the ROC curves in Figure \ref{fig:ROC}.
Variations of our proposed method are bolded.}
\par\end{centering}
\end{table}

\begin{figure}
\begin{centering}
\includegraphics[width=0.5\columnwidth]{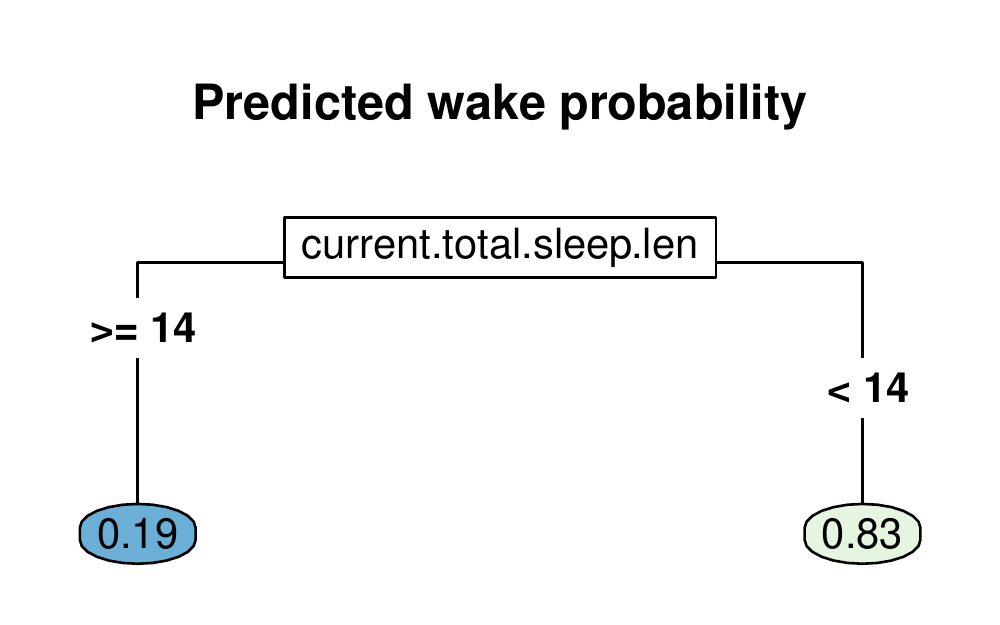}
\par\end{centering}
\caption{\label{fig:Webster's-rescoring-rules-fit} A version of Webster's
rescoring rules, calibrated to the MESA dataset. Current, total sleep
length is measured in minutes.}
\end{figure}

As an additional illustration of what optimized versions of rescoring
rules could resemble, Figure \ref{fig:Webster's-rescoring-rules-fit}
shows a version of the tree in Figure \ref{fig:Webster's-rescoring-rules}
trained on all 1685 participants. The input to the tree is a vector
of rescoring features $r(\mathbf{W},t)$, where $\mathbf{W}$ is the
binarized output of a moving window, logistic regression model, with
a window of $[-10,5]$. Here, the rescoring operation is simplified
to a single rule: any bout of predicted sleep lasting less than 14
minutes is rescored as wake.

\section{Discussion}

We have demonstrated how rescoring rules can be optimized for any
population of interest by reframing the rules in terms of epoch-level
features. In tests with the MESA dataset, we find our procedure to
produce accuracy comparable to certain configurations of LSTM networks.
That said, the improvement achieved by optimizing rescoring rules
is less noticeable when the initial moving window model has a wide
window. This limitation is intuitive, as some of the long-term information
otherwise gained from the rescoring rules is already attained from
the larger window. The implication of our applied analysis is that
larger windows, and optimization steps, should both be explored whenever
Webster's rescoring rules are used. This includes performance tests
where rescoring rules are used as a benchmark representing simple,
interpretable methods.

Our work opens up several avenues for future sleep-wake classification
methods. One simple extension would be to include other wearable device
measurements, such as heart rate, in the moving window model. Another
extension would be to explore other types of machine learning models
fit to the epoch-level features in Section \ref{sec:rescore-features}.
Digging more deeply, while we have held the formulas of our epoch-level
features ($l(\mathbf{W},t),n(\mathbf{W},t)$) fixed (see Appendix
\ref{sec:Recursive-definitions-for}), it could be fruitful to explore
versions of these features with trainable parameters. Rescoring features
could also be implemented in other neural network architectures. For
example, these transformations could be separately applied to the
output of several convolutional filters, or additional layers could
be stacked between the rescoring features $r(\mathbf{W},t)$ and the
final predictions (See Figure \ref{fig:structure}). 

An important caveat is that any joint optimization approach complicates
the interpretation of the rescoring rules, as their input is no longer
an estimated wake probability. This limitation is also true of our
Keras implementation (``rescore-NN''), but becomes especially problematic
in more complex network architectures. Even so, rescoring features
may still be helpful in regularizing prediction pipelines.

Our rescoring approach also immediately generalizes to multi-state
classification problems, such as general activity classification (e.g.,
sitting, walking, or stair climbing) or sleep stage classification.
For each state $k$, we can define epoch-level features such as ``the
time between now and the most recent (or closest upcoming) bout of
time in state $k$,'' using the same techniques as above (see Appendix
\ref{sec:Recursive-definitions-for}). In this way, rescoring features
could provide a promising means of incorporating long-term information
into general temporal modeling problems.

\section*{Acknowledgements}

The Multi-Ethnic Study of Atherosclerosis (MESA) Sleep Ancillary study
was funded by NIH-NHLBI Association of Sleep Disorders with Cardiovascular
Health Across Ethnic Groups (RO1 HL098433). MESA is supported by NHLBI
funded contracts HHSN268201500003I, N01-HC-95159, N01-HC-95160, N01-HC-95161,
N01-HC-95162, N01-HC-95163, N01-HC-95164, N01-HC-95165, N01-HC-95166,
N01-HC-95167, N01-HC-95168 and N01-HC-95169 from the National Heart,
Lung, and Blood Institute, and by cooperative agreements UL1-TR-000040,
UL1-TR-001079, and UL1-TR-001420 funded by NCATS. The National Sleep
Research Resource was supported by the National Heart, Lung, and Blood
Institute (R24 HL114473, 75N92019R002).

\bibliographystyle{apalike}
\bibliography{Paperpile-Websters-Rescoring}

\appendix

\section{Recursive definitions for $n(\mathbf{Y},t)$ and $l(\mathbf{Y},t)$\label{sec:Recursive-definitions-for}}

\textcolor{black}{Let $\epsilon$ be the length of an epoch. For $t>1$},
let
\begin{equation}
\text{last}_{\text{lag}}^{\text{wake}}(\mathbf{Y},t):=\left(1-Y_{t}\right)\left\{ \text{last}_{\text{lag}}^{\text{wake}}(\mathbf{Y},t-1)+\epsilon\right\} ,\label{eq:lwl}
\end{equation}
\[
\text{last}_{\text{lag}}^{\text{sleep}}(\mathbf{Y},t):=Y_{t}\left\{ \text{last}_{\text{lag}}^{\text{sleep}}(\mathbf{Y},t-1)+\epsilon\right\} ,
\]
\begin{align*}
\text{last}_{\text{len}}^{\text{wake}}(\mathbf{Y},t) & :=(1-Y_{t})\,\text{last}_{\text{len}}^{\text{wake}}(\mathbf{Y},t-1)+Y_{t}\,\text{last}_{\text{lag}}^{\text{sleep}}(\mathbf{Y},t)\\
 & =(1-Y_{t})\,\text{last}_{\text{len}}^{\text{wake}}(\mathbf{Y},t-1)+Y_{t}^{2}\,\left\{ \text{last}_{\text{lag}}^{\text{sleep}}(\mathbf{Y},t-1)+\epsilon\right\} \\
 & =(1-Y_{t})\,\text{last}_{\text{len}}^{\text{wake}}(\mathbf{Y},t-1)+Y_{t}\,\left\{ \text{last}_{\text{lag}}^{\text{sleep}}(\mathbf{Y},t-1)+\epsilon\right\} ,
\end{align*}
and
\begin{align}
\text{last}_{\text{len}}^{\text{sleep}}(\mathbf{Y},t) & :=Y_{t}\,\text{last}_{\text{len}}^{\text{sleep}}(\mathbf{Y},t-1)+(1-Y_{t})\,\left\{ \text{last}_{\text{lag}}^{\text{wake}}(\mathbf{Y},t-1)+\epsilon\right\} .\label{eq:lsl}
\end{align}

Following the same logic, for $t<T$, let

\[
\text{next}_{\text{lag}}^{\text{wake}}(\mathbf{Y},t):=(1-Y_{t})\left\{ \text{next}_{\text{lag}}^{\text{wake}}(\mathbf{Y},t+1)+\epsilon\right\} ,
\]
\[
\text{next}_{\text{lag}}^{\text{sleep}}(\mathbf{Y},t):=Y_{t}\left\{ \text{next}_{\text{lag}}^{\text{sleep}}(\mathbf{Y},t+1)+\epsilon\right\} ,
\]

\[
\text{next}_{\text{len}}^{\text{wake}}(\mathbf{Y},t):=(1-Y_{t})\,\text{\text{next}}_{\text{len}}^{\text{wake}}(\mathbf{Y},t+1)+Y_{t}\,\left\{ \text{\text{next}}_{\text{lag}}^{\text{sleep}}(\mathbf{Y},t+1)+\epsilon\right\} ,
\]
and
\[
\text{next}_{\text{len}}^{\text{sleep}}(\mathbf{Y},t):=Y_{t}\,\text{\text{next}}_{\text{len}}^{\text{sleep}}(\mathbf{Y},t+1)+(1-Y_{t})\,\left\{ \text{\text{next}}_{\text{lag}}^{\text{wake}}(\mathbf{Y},t+1)+\epsilon\right\} .
\]

It is fairly straightforward to generalize these features to the multi-state
setting. For each state $k$, we simply replace $Y_{t}$ with the
indicator $1(Y_{t}=k)$. The four ``wake'' features above then describe
bouts of time in state $k$, and the four ``sleep'' features describe
bouts of time \emph{not} spent in state $k$. In this way, the multi-state
setting generates approximately twice as many features per state as
the binary setting, since we additionally keep track of bouts of time
spent in any state but state $k$. These extra states are required
for our proposed computation of bout length (e.g., Eq \ref{eq:lsl}). 

Alternatively, these extra states can be removed if we instead use
a min operation when computing bout length. That is, we can set
\[
\text{last}_{\text{len}}^{k}(\mathbf{Y},t):=\left\{ 1-1(Y_{t}=k)\right\} \,\text{last}_{\text{len}}^{k}(\mathbf{Y},t-1)+1(Y_{t}=k)\,\min_{k'\neq k}\left\{ \text{last}_{\text{lag}}^{k'}(\mathbf{Y},t-1)+\epsilon\right\} 
\]
to be the length of the most recent bout in state $k$, and set 
\[
\text{last}_{\text{lag}}^{k}(\mathbf{Y},t):=\left\{ 1-1(Y_{t}=k)\right\} \left\{ \text{last}_{\text{lag}}^{k}(\mathbf{Y},t-1)+\epsilon\right\} 
\]
to be the time since that bout.

\section{Vectorized versions of rescoring features}

The four formulas in Eqs (\ref{eq:lwl})-(\ref{eq:lsl}) can also
be vectorized as follows.
\begin{align}
\left[\begin{array}{c}
\text{last}_{\text{lag}}^{\text{wake}}(\mathbf{Y},t)\\
\text{last}_{\text{lag}}^{\text{sleep}}(\mathbf{Y},t)\\
\text{last}_{\text{len}}^{\text{wake}}(\mathbf{Y},t)\\
\text{last}_{\text{len}}^{\text{sleep}}(\mathbf{Y},t)
\end{array}\right] & =\left[\begin{array}{c}
\epsilon\\
0\\
0\\
\epsilon
\end{array}\right]+\left[\begin{array}{c}
-\epsilon\\
\epsilon\\
\epsilon\\
-\epsilon
\end{array}\right]Y_{t}\label{eq:vec-line}\\
 & \hspace{1em}+\left(\left[\begin{array}{cccc}
1 & 0 & 0 & 0\\
0 & 0 & 0 & 0\\
0 & 0 & 1 & 0\\
1 & 0 & 0 & 0
\end{array}\right]+\left[\begin{array}{cccc}
-1 & 0 & 0 & 0\\
0 & 1 & 0 & 0\\
0 & 1 & -1 & 0\\
-1 & 0 & 0 & 1
\end{array}\right]Y_{t}\right)\left[\begin{array}{c}
\text{last}_{\text{lag}}^{\text{wake}}(\mathbf{Y},t-1)\\
\text{last}_{\text{lag}}^{\text{sleep}}(\mathbf{Y},t-1)\\
\text{last}_{\text{len}}^{\text{wake}}(\mathbf{Y},t-1)\\
\text{last}_{\text{len}}^{\text{sleep}}(\mathbf{Y},t-1)
\end{array}\right].\label{eq:mat-line}
\end{align}

Or, in more compact notation,

\[
l(\mathbf{Y},t)=\mathbf{v}_{0}+Y_{t}\mathbf{v}_{1}+(\mathbf{M}_{0}+Y_{t}\mathbf{M}_{1})l(\mathbf{Y},t-1),
\]
where $\mathbf{v}_{0}$ and $\mathbf{v}_{1}$ are the two vectors
on the right-hand side of Line (\ref{eq:vec-line}), and $\mathbf{M}_{0}$
and $\mathbf{M}_{1}$ are the two matrices in Line (\ref{eq:mat-line}).
Similarly, 
\[
n(\mathbf{Y},t)=\mathbf{v}_{0}+Y_{t}\mathbf{v}_{1}+(\mathbf{M}_{0}+Y_{t}\mathbf{M}_{1})n(\mathbf{Y},t+1).
\]

\section{\label{sec:Expectation-of}Conditional expectation of $l(\mathbf{Y},t)$
and $n(\mathbf{Y},t)$}

In this section we show the claim from Section \ref{sec:Improving-Webster's-rescoring}
that if $(Y_{1},\dots,Y_{T})$ are independently distributed Bernoulli
variables, given\textbf{ $\mathbf{X}$}, with $P(Y_{t}=1|\mathbf{X})=\hat{\pi}_{t}$,
then $E\left[l(\mathbf{Y},t)|\mathbf{X}\right]=l(\hat{\pi},t)$ and
$E\left[n(\mathbf{Y},t)|\mathbf{X}\right]=n(\hat{\pi},t)$. 

For $l(\mathbf{Y},t)$ we show this by induction. The base case holds
by definition, since $E\left[l(\mathbf{Y},1)|\mathbf{X}\right]=b_{1}=l(\hat{\pi},1)$.
For the inductive step, note that $l(\mathbf{Y},t)$ depends only
on $(Y_{1},\dots,Y_{t})$. If $E\left[l(\mathbf{Y},t-1)|\mathbf{X}\right]=l(\hat{\pi},t-1)$,
then the vector representations in the previous section tell us that
\begin{align}
 & E\left[l(\mathbf{Y},t)|\mathbf{X}\right]\nonumber \\
 & =E\left[E\left\{ l(\mathbf{Y},t)|Y_{1},\dots,Y_{t-1},\mathbf{X}\right\} \,|\,\mathbf{X}\right]\nonumber \\
 & =E\left[E\left\{ \mathbf{v}_{0}+Y_{t}\mathbf{v}_{1}+(\mathbf{M}_{0}+Y_{t}\mathbf{M}_{1})l(\mathbf{Y},t-1)\hspace{1em}|\hspace{1em}Y_{1},\dots,Y_{t-1},\mathbf{X},l(\mathbf{Y},t-1)\right\} \,|\,\mathbf{X}\right]\nonumber \\
 & =E\left[E\left\{ \mathbf{v}_{0}+Y_{t}\mathbf{v}_{1}+(\mathbf{M}_{0}+Y_{t}\mathbf{M}_{1})l(\mathbf{Y},t-1)\hspace{1em}|\hspace{1em}\mathbf{X},l(\mathbf{Y},t-1)\right\} \,|\,\mathbf{X}\right]\nonumber \\
 & =E\left[\mathbf{v}_{0}+E\left\{ Y_{t}|l(\mathbf{Y},t-1),\mathbf{X}\right\} \mathbf{v}_{1}+(\mathbf{M}_{0}+E\left\{ Y_{t}|l(\mathbf{Y},t-1),\mathbf{X}\right\} \mathbf{M}_{1})l(\mathbf{Y},t-1)\,|\,\mathbf{X}\right]\nonumber \\
 & =E\left[\mathbf{v}_{0}+E\left\{ Y_{t}|\mathbf{X}\right\} \mathbf{v}_{1}+(\mathbf{M}_{0}+E\left\{ Y_{t}|\mathbf{X}\right\} \mathbf{M}_{1})l(\mathbf{Y},t-1)\,|\,\mathbf{X}\right]\label{eq:cond-ind}\\
 & =E\left[\mathbf{v}_{0}+\hat{\pi}_{t}\mathbf{v}_{1}+(\mathbf{M}_{0}+\hat{\pi}_{t}\mathbf{M}_{1})l(\mathbf{Y},t-1)\,|\,\mathbf{X}\right]\nonumber \\
 & =\mathbf{v}_{0}+\hat{\pi}_{t}\mathbf{v}_{1}+(\mathbf{M}_{0}+\hat{\pi}_{t}\mathbf{M}_{1})E\left[l(\mathbf{Y},t-1)\,|\,\mathbf{X}\right]\nonumber \\
 & =\mathbf{v}_{0}+\hat{\pi}_{t}\mathbf{v}_{1}+(\mathbf{M}_{0}+\hat{\pi}_{t}\mathbf{M}_{1})l(\hat{\pi},t-1)\nonumber \\
 & =l(\hat{\pi},t).\nonumber 
\end{align}
Above, Line (\ref{eq:cond-ind}) comes from our assumption of conditional
independence. 

The same steps can be used to show that $E\left[n(\mathbf{Y},t)|\mathbf{X}\right]=n(\hat{\pi},t)$.
\end{document}